\documentclass[letterpaper, 10 pt, conference]{ieeeconf}  % Comment this line out if you need a4paper

\IEEEoverridecommandlockouts                              % This command is only needed if 
\usepackage{graphicx} % for pdf, bitmapped graphics files
\usepackage{amsmath}
\usepackage{amssymb}
\usepackage{cite}
\usepackage{booktabs}
\usepackage{multirow}
\usepackage{xcolor}

\newcommand{\methodname}{%Information Bottlenecking via Double Transformation
Sim2Real via Bottlenecked Latent Reconstruction}

\newcommand{\methodabbr}{SBLR}

\title{\LARGE \bf
Tactile Sim2Real without Tactile Simulation via Bottlenecked Latent Reconstruction
}
\author{Fan Yang$^{1}$, Youngsun Wi$^{2}$, Jinhao Yu$^{1}$, Nima Fazeli$^{1}$, Dmitry Berenson$^{1}$% <-this % stops a space
\thanks{$^{1}$F. Yang, J. Yu, N. Fazeli, and D. Berenson are with the University of Michigan, Ann Arbor, MI, USA.}
\thanks{$^{2}$Y. Wi is with Google DeepMind. This work was initiated prior to and conducted independently of Google DeepMind.}
}

\begin{document}

\maketitle
\thispagestyle{empty}
\pagestyle{empty}

%%%%%%%%%%%%%%%%%%%%%%%%%%%%%%%%%%%%%%%%%%%%%%%%%%%%%%%%%%%%%%%%%%%%%%%%%%%%%%%%

\begin{abstract}
% Sim-to-real transfer for robot manipulation is often hampered by the difficulty of faithfully simulating real sensor readings, which requires substantial domain expertise and is further limited by computational approximations that can corrupt the simulated signals. 
Robot sensor designs, particularly tactile sensors, are highly diverse and evolve rapidly. Modeling each sensor in simulation demands substantial domain expertise and computational approximations can degrade the fidelity of the simulated signals.
We propose \methodname~(\methodabbr), a framework that avoids sensor-specific simulation entirely by (1)~training policies on a simulator-native oracle sensor that is easy to construct without modeling any particular sensor (e.g. we use a point-cloud and finger-tip forces as a tactile oracle), and (2)~aligning real sensor latent embeddings to those of the oracle sensor at inference time. 
% \reupdate{We instantiate our framework on tactile sensing with Point Force Composition (PFC)---fingertip and object full point clouds with contact forces---as the oracle sensor reading.} 
% Policy training proceeds in two stages: in stage one, the policy learns the task using oracle sensor latent embeddings directly; in stage two, a novel double-transformation architecture exposes the policy to the information loss induced by the real sensor through a forward-and-back transformation, adapting it to the gap between the information-rich oracle sensor and the real sensor. 
Policy training proceeds in two-stage: the policy first learns from the oracle sensor latents, then undergoes a lightweight second-stage adaptation with bottlenecked latent reconstruction to account for the information loss expected when using the real sensor instead of the oracle.
The alignment between oracle and real sensor is learned from unpaired random-play data collected in both simulation and the real world, using rectified-flow-based transformation networks trained on nearest-neighbor pseudo-pairs.
Simulation experiments on three contact-rich tasks show that \methodabbr\
matches or approaches the performance of an oracle with direct access to tactile simulation. Hardware experiments on Peg Insertion and Gear Meshing with GelSight Mini and DIGIT sensors demonstrate 85--97.5\% zero-shot success without requiring any sensor-specific modeling or calibration, outperforming a physics-based tactile simulation baseline by 7.5--15\%.
\looseness=-1
\end{abstract}

\section{Introduction}

Learning robot manipulation policies in simulation and transferring them zero-shot to the real world is a promising approach for contact-rich manipulation tasks, benefiting from advantages such as massive parallelization and privileged information that is difficult to obtain in the real world ~\cite{qi2023hand, qi2023general, lum2025crossing, yin2025dexteritygen}. 
The performance of sim-to-real transfer depends heavily on the fidelity of the simulated sensors used for training the policy, yet some sensing modalities, such as tactile sensors, which are valuable for manipulation, are difficult to reproduce faithfully in simulation.
% This creates a major difficulty for sim-to-real transfer: faithfully simulating a sensor's real-world readings typically requires substantial domain expertise, both in computational simulation methods and the physics of the sensor. 

Faithfully simulating a sensor's real-world readings typically requires substantial domain expertise in both computational simulation methods and the underlying sensor physics. 
This is especially challenging for tactile sensing, where sensor designs are highly diverse~\cite{yuan2017gelsight,lambeta2020digit, tomo2017covering}, and each design typically requires its own physics-based model. % TODO: cite additional sensor families
Additionally, tactile sensor designs evolve rapidly, often outpacing the engineering effort needed to build and validate the sensor simulations.
This modeling and validation process raises a high barrier to adopting such sensors in simulation.
% Even sensors of the same design can differ in manufacturing and produce slightly different readings, requiring careful calibration of sensor parameters to the specific unit used in the real world~\cite{si2022taxim, akinola2025tacsl, li2026taccel}.
% This modeling and calibration process is time-consuming and raises a high barrier to adopting such sensors in simulation.
Even then, simulated sensor readings rarely match their real-world counterparts perfectly, so extensive domain randomization~\cite{tobin2017domain} is typically required to make the trained policy robust to the remaining sim-to-real gap from the sensor modality~\cite{han2025zero, akinola2025tacsl, pan2026beyond, he2025viral}. However, domain randomization can wash out the fine-grained sensory details that matter for manipulation~\cite{james2018sim}, degrading performance on high-precision tasks.
% \looseness=-1

\begin{figure}[t]
    \centering
    \includegraphics[width=\columnwidth]{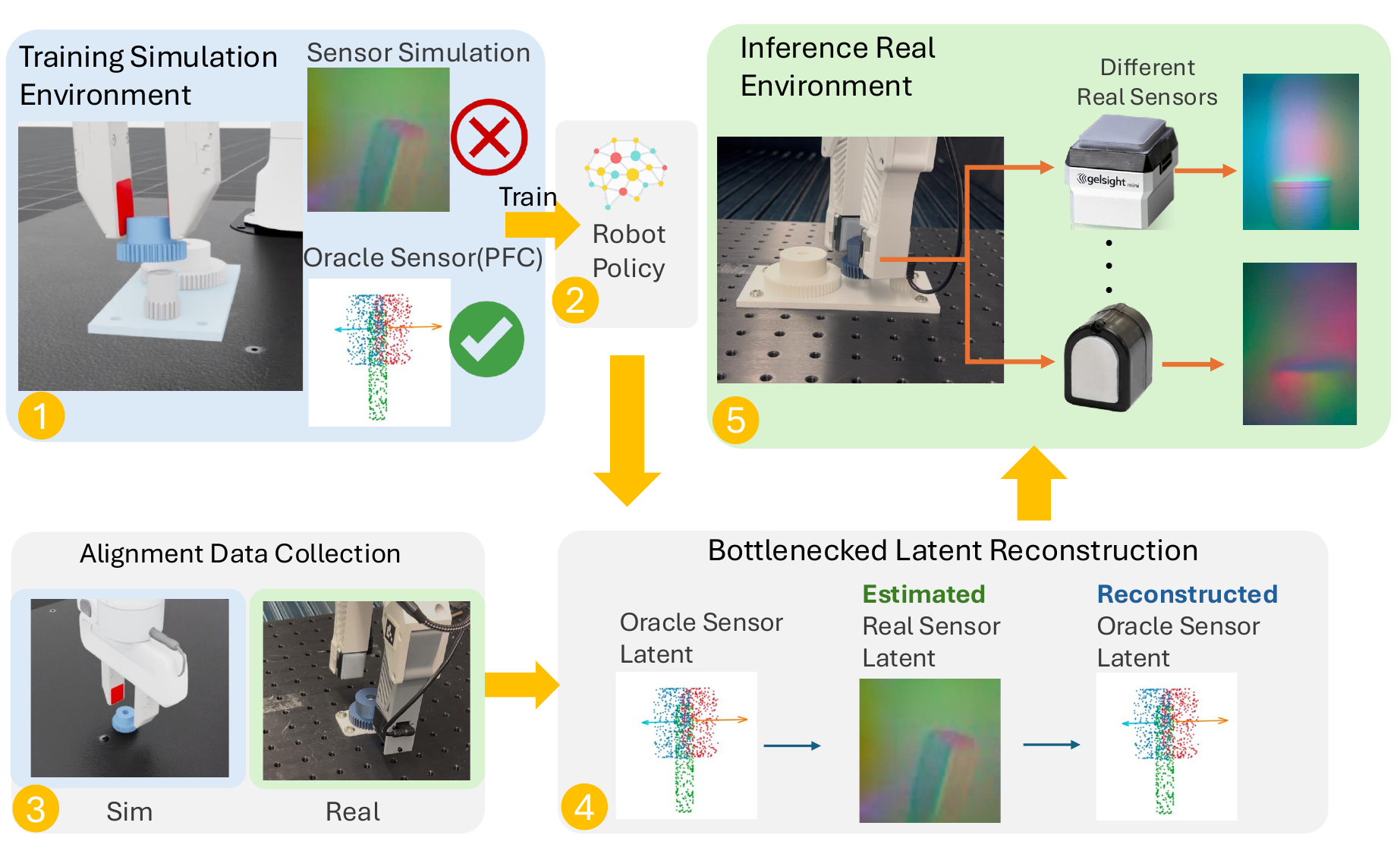}
    \vspace{-0.7cm}
    \caption{
    Instead of training on physics-based sensor simulation, we train RL policies on an oracle sensor available in most simulators without additional sensor setup.
    A bottlenecked latent reconstruction stage then adapts the policy by reconstructing oracle latents through the information available from real sensors, enabling zero-shot deployment across different sensors (e.g., GelSight Mini and DIGIT) in the real world.}
    \label{fig:pull}
    \vspace{-0.7cm}
\end{figure}
% Thus, successful sim-to-real transfer requires a careful consideration of the sensor design and the sensor readings***CITE***. This includes using model-based methods to simulate the sensor readings and modify the simulator accordingly to incorporate the tactile sensor modality ***CITE***, and using data-driven methods to calibrate the sensor readings to the real world ***CITE***. These steps generally require intensive human efforts and domain expertise, which limits the scalability of sim-to-real manipulation with tactile sensing.

% To address the challenge of modeling hard-to-simulate sensors for sim-to-real transfer, we propose \methodname\ (Fig.~\ref{fig:pull}), a framework whose core idea is to (1)~design an oracle sensor \update{that acts as ideal sensing by providing at least as much information as the real sensor and} is readily available in most simulators without additional sensor setup, \update{(2)~collect unpaired random-play data in both simulation and the real world to train transformation networks between the latent of the oracle and real sensor readings,} (3)~train robot manipulation policies on this oracle sensor without using a sensor-specific simulation, \update{(4)~adapt the policy using a novel double-transformation method to the information gap with the real sensor,} and (5)~align real sensor embeddings to the oracle embedings at inference time. 
To address the challenge of modeling hard-to-simulate sensors, e.g., tactile sensors, for sim-to-real transfer, we propose \methodname\ (Fig.~\ref{fig:pull}), a method that adapts oracle-trained policies to the information available from real sensors.
Concretely, instead of simulating any particular real sensor, we train policies on a \textbf{simulator-native} oracle sensor that provides at least as much information as the real tactile sensor, and align real sensor latents to oracle latents at inference using a novel information bottlenecking method.
The alignment module used at both inference and bottlenecked reconstruction is trained on unpaired random-play data collected in both simulation and the real world.
With this alignment, a policy can be trained in simulation without knowing which real sensor will be used at deployment. In contrast, prior methods typically assume a specific sensor~\cite{huang2025tactile, zhang2026vtla, bi2025vla}, and require costly retraining for each new sensor setup. 
\looseness=-1

However, alignment alone does not fully close the modality gap: the oracle sensor typically conveys more information than real sensors, so a perfect transformation from real sensor latent embeddings (real latents) to the oracle sensor latent embeddings (oracle latents) is impossible, and a policy trained only on oracle latents can degrade when it receives these imperfect reconstructions at inference. 
We address this in a second RL stage that uses bottlenecked latent reconstruction to close this information gap. Concretely,
we use two Rectified Flow (RF)~alignment networks in a double transformation architecture: the first RF maps the oracle latents to estimated real latents, removing information unavailable from the real sensor, and then the second RF maps those estimated real latents back to the oracle, approximating the distribution encountered at inference.
% The first RF removes information that is present in the proxy representation but unavailable from the real sensor; the second RF maps the predicted real readings back to the proxy, approximating the distribution encountered at inference time. Both RF networks are frozen during this second-stage RL training.
% The second stage is initialized from the first-stage policy and only needs to adapt to the information gap, rather than learning the manipulation strategy from scratch. As a result, it is much more lightweight, requiring far fewer training epochs than the first stage.

While the framework is agnostic to the underlying sensor modality, in this paper we instantiate it only on tactile sensing.
We focus on tactile sensors because sensor designs are highly diverse and accurate tactile simulation remains a major difficulty for sim-to-real transfer.
The oracle sensor we use is Point Force Composition (PFC): fingertip and object \text{full} point clouds with contact forces.
% Our tactile oracle sensor design is motivated by the observation that, in many manipulation tasks, tactile sensors mainly provide (1)~the pose and local geometry of the manipulated object in the robot gripper frame and (2)~the contact forces acting on the object. Based on this observation, we propose Point Force Composition (PFC), a tactile representation that can be easily set up for policy training in simulation. PFC encodes these two cues explicitly: (1) a full point cloud of both fingertips and the manipulated object for object pose and local geometry relative to the gripper, while (2) contact forces at each fingertip for force information.

Our contributions are summarized as follows:
\begin{itemize}
    \item We propose \methodname~(\methodabbr), a framework that trains policies on a simulator-ready oracle sensor and aligns real sensor latents to oracle latents at inference time via unpaired random-play data, removing the need for physics-based sensor simulation.
    \item We introduce a bottlenecked latent reconstruction architecture that adapts the RL policy to the information gap between oracle latents and real sensor latents through a lightweight second stage of RL training.
    \item We instantiate the framework on tactile sensing with PFC as the oracle sensor, and demonstrate zero-shot sim-to-real transfer with real tactile sensors.
\end{itemize}
We evaluate our method on hardware contact-rich manipulation tasks with two different tactile sensors, achieving 85--97.5\% zero-shot success without sensor-specific simulation. We also outperform a physics-based tactile simulation baseline. We further report simulation experiments that isolate the sensor modality gap from the dynamics gap, providing a clearer assessment of our method.

\section{Related Work}
\subsection{Unified Tactile Representation}
Many recent works~\cite{higuera2024sparsh, zhao2024transferable, feng2025anytouch, yang2024binding} focus on learning a unified tactile representation that transfers across various tactile sensors for manipulation.
Others~\cite{cheng2025omnivtla, yuan2026ftp} go further and train Vision-Language-Action (VLA) models on such unified representations, enabling training and inference with heterogeneous sensors.
Our oracle sensor may appear similar in that it also supports multiple real-world sensors, but it differs in two key ways.
First, unified representations extract \emph{shared} information across sensors, whereas our oracle sensor is designed to convey \emph{more} information than any real sensor, and our algorithm has an additional step to adapt to this information gap.
Second, these approaches act more like pretrained tactile encoders: they do not remove the need for sensor-specific tactile simulation when training policies for sim-to-real transfer.
\looseness=-1

\subsection{Physics-based Tactile Simulation}
Recent work has made substantial progress on physics-based tactile simulation.
Taxim~\cite{si2022taxim} proposes calibration methods for tactile image rendering, shadow processing, and tactile marker modeling.
FOTS~\cite{zhao2024fots} generates realistic lighting and shadows in a computationally efficient manner.
TacSL~\cite{akinola2025tacsl} and Taccel~\cite{li2026taccel} further scale tactile simulation to massive parallelization on GPU-accelerated simulators.
Hydroshear~\cite{dang2026hydroshear} models shear forces during contact, and Pan et al.~\cite{pan2026beyond} extend tactile simulation beyond visuotactile sensors to devices such as Xela~\cite{tomo2017covering}.
These methods have enabled successful sim-to-real transfer for RL policies.
However, tactile sensor designs remain highly diverse, and setting up a simulation for each specific design requires domain expertise and is prone to error. Our method seeks to avoid creating such simulations.%exactly the challenge we aim to address.
\subsection{Cross-Domain Representation Alignment}
Aligning latent representations across environments, e.g., sim and real, or human and robot, is an active research direction.
TactSpace~\cite{joarder2026tactspace} aligns tactile representations between simulation and the real world, closely related to our setting, but still requires a Finite Element Method (FEM)-based tactile sensor model in simulation and does not evaluate on robot manipulation tasks.
Rodriguez et al.~\cite{rodriguez2025cross} align different tactile sensors via depth information, but likewise do not evaluate on manipulation.
Cheng et al.~\cite{cheng2026generalizable} use optimal transport to align camera images across sim and real, and EgoMimic~\cite{kareer2024egomimic} aligns human egocentric videos with robot videos; both are limited to vision from cameras and assume that the sensors in both domains produce images. 
PTLD~\cite{chen2026ptld} distills object-state information into tactile observations, while we align tactile sensing with oracle contact geometry and forces. 
Closest to our work, TactAlign~\cite{wi2025tactalign} aligns tactile readings from human demonstrations with a tactile glove to robot demonstrations.
Unlike our work, however, TactAlign does not address tactile sensor simulation or training RL policies for sim-to-real transfer.
\looseness=-1

\section{Problem Statement}
We focus on the problem of training a policy in simulation without using physics-based modeling of any specific sensor. We instantiate this setup on tactile sensing. We evaluate performance by the task success rate under zero-shot real-world deployment.

In simulation, we assume access to an oracle sensor with its readings $\mathbf{p}$ that are readily available without sensor-specific simulation.
To align $\mathbf{p}$ with real tactile sensor readings $\mathbf{c}:= \{\mathbf{c}_i\}_{i}$, where $\mathbf{c}_i$ is the reading from the sensor on the $i$th finger and $\mathbf{c}$ is their concatenation, we also assume access to an unpaired alignment dataset of robot random-play data in simulation and in the real world. In this dataset, the robot randomly interacts with an object that remains fixed during collection and will also be used at inference, collecting contacts similar to those encountered during performance of the task.

The oracle sensor reading (oracle reading) $\mathbf{p}$ is Point Force Composition (PFC), which encodes the two cues tactile sensing typically provides in manipulation---object pose and local geometry relative to the gripper, and contact forces: $\mathbf{p}:= \big(\{\mathbf{X}_i, \mathbf{f}_i\}_{i}, \mathbf{X}_o\big)$, where $\mathbf{X}_o$  and $\mathbf{X}_i$ are \textbf{full} point clouds of the manipulated object and each robot fingertip, and $\mathbf{f}_i$ is the contact force at fingertip $i$.

States collected in simulation datasets are proprioceptions and oracle readings: $\{\mathbf{q}, \mathbf{p}\}$, where $\mathbf{q}$ is robot proprioception. Real-world states include tactile readings rather than oracle readings: $\{\mathbf{q}, \mathbf{c}\}$.
Because real-world data collection is generally more expensive, we assume a limited size of a real dataset and a much larger simulation dataset for alignment.

Our objective is to train a policy $\pi_{\theta}(\mathbf{o}, \mathbf{p})$ in simulation on the oracle readings and deploy it zero-shot in the real world with high success rate, where  $\mathbf{c}$ is available but not $\mathbf{p}$. Here, $\mathbf{o}$ denotes non-tactile observations such as proprioception and the previous action.
% We aim to match the performance of an oracle policy $\hat{\pi}_{\theta}(\mathbf{o}, \mathbf{c})$ that is trained in simulation but with access to a model of real sensor readings.

\section{Method}
\begin{figure*}[t]
    \centering
    \includegraphics[width=0.9\textwidth]{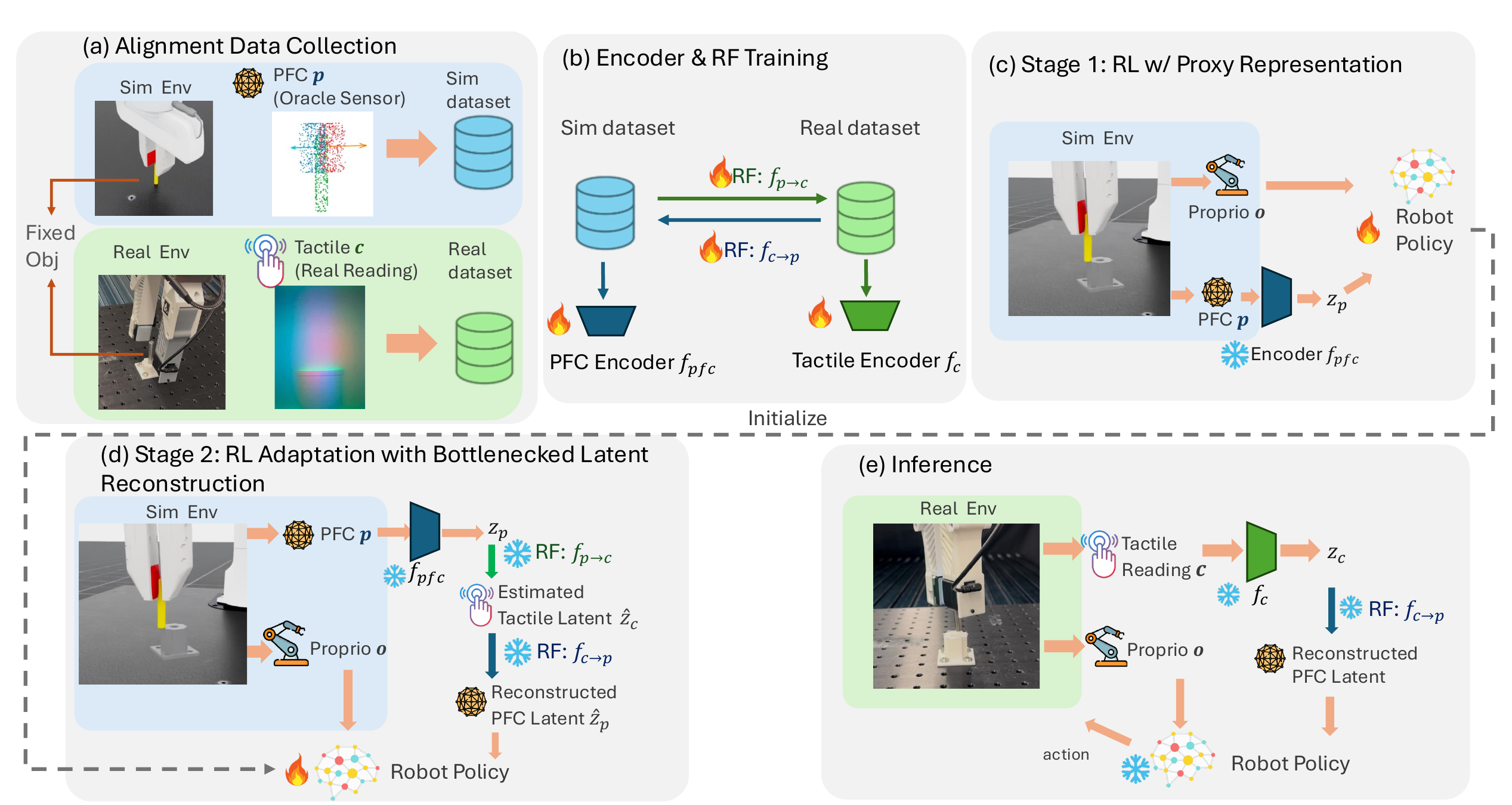}
    \vspace{-0.4cm}
    \caption{\textbf{Overview of \methodname.}
    Unpaired PFC (sim) and tactile (real) data is used to train bidirectional rectified-flow converters $f_{p\rightarrow c}$ and $f_{c\rightarrow p}$.
    Stage~1 learns a base policy on PFC without sensor-specific simulation; a lightweight stage~2 adapts it via bottlenecked latent reconstruction.
    At deployment, $f_{c\rightarrow p}$ maps real tactile readings to PFC for the adapted RL policy.}
    \label{fig:method}
    \vspace{-0.6cm}
\end{figure*}

Our method trains an RL policy on oracle sensor latents in simulation, adapts it via bottlenecked latent reconstruction to the real-sensor information gap, and zero-shot deploys the policy in the real world without collecting real-world data directly tied to the primary task (e.g. without real-world RL or expert demonstrations).
Our framework consists of four steps: (1)~collect unpaired random-play alignment data in simulation and the real world (Fig.~\ref{fig:method}(a)); (2)~pretrain encoders for the oracle and real sensor readings, and train two transformation networks $f_{p \rightarrow c}$ and $f_{c \rightarrow p}$ that map the oracle latents to real sensor latents and vice versa (Fig.~\ref{fig:method}(b)); (3)~train an RL policy $\pi_{\theta}(\mathbf{o}, \mathbf{p})$ in simulation on the oracle readings, yielding a base policy with a high success rate (Fig.~\ref{fig:method}(c)); and (4)~adapt this policy with a second stage of RL training to the information gap between oracle and real sensor latents, using the frozen transformation networks from step (2) (Fig.~\ref{fig:method}(d)). Our main contributions are the RF-based transformation networks in Sec.~\ref{sec:rectified-flow-training} and how we address the information gap between the oracle and real sensor latents with bottlenecked latent reconstruction in Sec.~\ref{sec:rl-training-with-double-transformation}.
\subsection{Alignment Data Construction}
\label{sec:alignment-data-construction}
The goal of alignment data construction is to characterize the mapping between PFC and real tactile readings over the contact distribution expected during inference. Training the alignment module requires pairing PFC with real readings, but obtaining an exact ground-truth pairing would require reproducing identical contacts across simulation and reality, which is very challenging to enforce. We therefore collect unpaired random-play data in both domains and form pseudo-pairs with nearest-neighbor search.

Concretely, we fix the manipulated object at a known pose and command the robot to make automated random probing contacts in both simulation and the real world. Because the object pose is shared across domains, similar proprioceptive transitions indicate similar robot--object contacts, so we form pseudo-pairs by nearest-neighbor search over $(\mathbf{q}_{t-1}, \mathbf{q}_t)$ and, for each real sample, retain the top-$k$ nearest simulation samples within a distance threshold.
Full data-collection and pairing details are shown in Appendix~\ref{app:alignment-data}.
\subsection{PFC and Image Encoder Training}
Raw point clouds and images are high-dimensional and difficult to process directly; cross-modal transformation, which will be used in Sec.~\ref{sec:rectified-flow-training}, is even more difficult if performed in the raw point cloud or image space. 
Therefore, we pretrain two encoders for PFC and tactile readings respectively that embed both modalities into a lower-dimensional latent space in a self-supervised manner. For the point-cloud component of PFC, we use a Point Masked Autoencoder (Point-MAE)~\cite{pang2023masked}.
Point-MAE divides the point cloud into patches and produces a latent for each. Every patch latent aggregates information from the full point cloud, and we average these latents to obtain a global latent representation $\mathbf{z}_x = f_x(\mathbf{X})$.
For the force component, we train a separate autoencoder with a reconstruction loss. 
The encoding process can be written as $\mathbf{z}_f = f_f(\mathbf{f})$. The PFC latent is the concatenation of both: $\mathbf{z}_{p} = \{ \mathbf{z}_x, \mathbf{z}_f \}$. 
The encoding process of PFC can be written as $\mathbf{z}_{p}=f_{pfc}(\mathbf{p})$.

For tactile readings, the choice of encoding method depends on the sensor modality; our framework does not require a particular architecture. In our experiments, we use visuotactile sensors and encode readings from each sensor with a Masked Autoencoder (MAE)~\cite{he2022masked}, $\mathbf{z}_{c_i} = f_c(\mathbf{c}_i)$. 
% \update{Although pretrained encoders such as Sparsh~\cite{higuera2024sparsh} are also compatible with our framework, we train an MAE on our play dataset, whose contact distribution matches that at inference.}
Similar to Point-MAE, $\mathbf{z}_{c_i}$ is the average of per-patch latents. 
The multi-sensor tactile latent is the concatenation of these per-sensor latents: $\mathbf{z}_c = [\mathbf{z}_{c_i}]_i$.

Because the alignment data (Sec.~\ref{sec:alignment-data-construction}) is intended to cover the contact distribution at inference time, we train the PFC encoder $f_{pfc}$ only on the simulation dataset and the tactile encoder $f_c$ only on the real-world dataset.
\subsection{Rectified-Flow-Based transformation Networks}
\label{sec:rectified-flow-training}
We train transformation networks $f_{p \rightarrow c}$ and $f_{c \rightarrow p}$ with Rectified Flow (RF)~\cite{liu2022flow} to map oracle latents (PFC) to corresponding tactile sensor latents and vice versa. 
We use RF rather than a deterministic supervised map because RF can model multimodal distributions. This is useful under information asymmetry between oracle and real sensor latents: for example, if the real sensor is less informative than PFC, a single tactile sensor latent may correspond to multiple oracle latents.
Below we only describe $f_{p \rightarrow c}$; $f_{c \rightarrow p}$ is symmetric by swapping the variables.

Given the pseudo-pairs from Sec.~\ref{sec:pseudo-pairing}, we train a velocity field $\mathbf{v}_{\psi}(\mathbf{z}_{t}, t, \mathbf{z}_{p})$ conditioned on the PFC latent $\mathbf{z}_{p}$ that maps samples from a source Gaussian noise sample $\mathbf{z}_{0} \sim \mathcal{N}(0, I)$ to the paired tactile latent $\mathbf{z}_{c}$, where $t \in [0, 1]$ is normalized flow time and $\mathbf{z}_{t}$ is the flow state at $t$. During training, we sample $\mathbf{z}_{t}$ from the straight-line bridge
\begin{equation}
\mathbf{z}_{t} = (1-t)\mathbf{z}_{0} + t \mathbf{z}_{c}.
\end{equation}
Assuming a constant flow velocity transitioning from $\mathbf{z}_0$ to $\mathbf{z}_{c}$, the training loss is
\begin{equation}
\mathcal{L}_{\mathrm{RF}} = \mathbb{E}_{t,\mathbf{z}_{0},\mathbf{z}_{c}}\left[\left\|\mathbf{v}_{\psi}(\mathbf{z}_{t}, t, \mathbf{z}_{p}) - (\mathbf{z}_{c} - \mathbf{z}_{0})\right\|_{2}^{2}\right].
\end{equation}
At inference, we solve the ordinary differential equation $d\mathbf{z}_{t}/dt = \mathbf{v}_{\psi}(\mathbf{z}_{t}, t, \mathbf{z}_{p})$ by integrating the velocity field. The predicted tactile latent is
\begin{equation}
\mathbf{z}_{c} = \mathbf{z}_{0} + \int_{0}^{1} \mathbf{v}_{\psi}(\mathbf{z}_{t}, t, \mathbf{z}_{p}) \, dt.
\label{eq:rf-inference}
\end{equation}
For notational convenience, we write the generation obtained by integrating Eq.~\ref{eq:rf-inference} as $\mathbf{z}_c \sim f_{p \rightarrow c}(\mathbf{z}_{p})$.

The transformation network conditions only on $\mathbf{z}_{p}$ and does not take robot proprioception $\mathbf{q}$ as input, even though additional conditioning could potentially improve transformation accuracy. We omit $\mathbf{q}$ because the network is trained on off-task play data, whose proprioceptive distribution differs from that of the main manipulation task.
\subsection{RL Training with PFC}
\label{sec:rl-training-with-pfc}
In the first stage of policy training, we train a PFC policy $\pi_{\theta}(\mathbf{o}, f_{pfc}(\mathbf{X}, \mathbf{f}))$ with PPO~\cite{schulman2017proximal}. This stage is agnostic to any real tactile sensor. The PFC encoder $f_{pfc}$ is frozen. The goal is to acquire basic task proficiency using information that real tactile sensors may eventually provide, such as object pose and contact forces.
At deployment, real tactile readings are mapped into the PFC latent space with $f_{c \rightarrow p}$ so that the same policy can be applied; see Sec.~\ref{sec:inference} for details.

We apply light randomization to PFC to prevent overfitting to idealized oracle observations and to suppress simulator-specific contact artifacts unavailable from real sensors.
% We also apply light randomization to PFC during training, for two reasons: (1)~even with a well-designed sensor, reconstructing a perfectly clean PFC from real readings is impossible; a policy trained only on clean PFC can overfit to idealized PFC and add unnecessary difficulties for Stage~2 adaptation in Sec.~\ref{sec:rl-training-with-double-transformation}; (2)~simulated contact forces are obtained from the physics model under simplifying assumptions (e.g., rigid contacts), and these assumptions introduce artifacts that are inaccurate and also potentially leak contact information unavailable from any real sensor.
This randomization is intentionally small and is not intended to match the noise level of the real sensor used in our experiments. Concretely, when rendering the point cloud, we add Gaussian noise to the object pose with standard deviation $0.5$\,mm in position and $0.29^{\circ}$ in rotation, and add Gaussian noise of $0.3$\,N to the contact force.

Real tactile sensors vary in design, however, and are not guaranteed to match the information content of PFC. For example, low-end visuotactile sensors may not provide high-precision contact forces, which can substantially degrade a policy that relies on accurate contact force observations. We therefore introduce a second RL stage in Sec.~\ref{sec:rl-training-with-double-transformation}.
\subsection{RL Training with Bottlenecked Latent Reconstruction}
\label{sec:rl-training-with-double-transformation}
Because PFC and real tactile readings may convey different information, a PFC policy is likely to be suboptimal if deployed directly in the real world.
We address this with a second RL training stage that adapts the policy to the information gap between PFC and tactile readings. The policy is initialized from the Stage~1 PFC policy of Sec.~\ref{sec:rl-training-with-pfc}.
We then use the two RF networks, $f_{p \rightarrow c}$ and $f_{c \rightarrow p}$:
at each time step, we first map the latent of the simulated PFC, $\mathbf{z}_{p}$, to corresponding real tactile readings $\hat{\mathbf{z}}_{c} = f_{p \rightarrow c}(\mathbf{z}_{p})$, rendering what the real-world tactile 
sensor readings would be for that contact, but without using a sensor simulation. 
However, because the policy expects $\mathbf{z}_{p}$ rather than $\mathbf{z}_{c}$, $\hat{\mathbf{z}}_{c}$ cannot be used directly for training.
We then map the generated readings back to PFC:
\looseness=-1
\begin{equation}
   \hat{\mathbf{z}}_{p} = f_{c \rightarrow p}(f_{p \rightarrow c}(\mathbf{z}_{p})). 
\end{equation}
Even with perfect transformation networks, the reconstructed latent $\hat{\mathbf{z}}_{p}$ will not necessarily match the original $\mathbf{z}_{p}$ whenever an information gap exists between PFC and real readings. 
For example, when one tactile latent corresponds to multiple PFC latents, reconstructing through $\hat{\mathbf{z}}_{c}$ may not recover the original $\mathbf{z}_{p}$.
Such differences are introduced by design to address the information gap: it exposes the policy to the information loss induced by the real sensor. 
In practice, perfect RF transformation is rarely attainable due to network approximation errors and pseudo-pairing errors, so bottlenecked latent reconstruction also exposes the policy to those RF alignment-network errors. 

Because $f_{c \rightarrow p}$ is also used at inference, bottlenecked reconstruction encourages the PFC observation distribution seen by the RL policy during stage 2 training to match that seen at deployment. Specifically, the policy is trained on $f_{c \rightarrow p}(\hat{\mathbf{z}}_{c})$ and evaluated at inference on $f_{c \rightarrow p}(f_c(\mathbf{c}))$, where $f_{c \rightarrow p}$ is applied to both scenarios; see Sec.~\ref{sec:inference}.
Both transformation networks and encoders remain frozen during this stage.

The two-stage RL procedure is not the only way to use the RF networks to address the modality gap between PFC and real sensor readings. 
Alternatively, one could train with bottlenecked latent reconstruction from scratch (omitting Stage~1), or train a policy directly on estimated real sensor latents, $\pi_{\theta}(\mathbf{o}, f_{p\rightarrow c}(\mathbf{z}_{p}))$, which can then be deployed on real tactile latents without an RF network at inference.
We use the two-stage design because adapting to a new sensor setup requires only the lightweight Stage~2 RL training, whereas the alternatives require retraining the entire policy, which can be extremely expensive for large models such as VLAs. In our experiments, adapting the policy to the real-world sensor in Stage~2 requires only 13.3\% of the training budget used for Stage~1.
\looseness=-1

\subsection{Inference}
\label{sec:inference}
At deployment, only real tactile readings $\mathbf{c}$ are available. Because the policy expects PFC latents, we use the conversion network $f_{c \rightarrow p}$ to map tactile latents into the PFC latent space:
\begin{equation}
\mathbf{a} = \pi_{\theta}(\mathbf{o}, f_{c \rightarrow p}(f_c(\mathbf{c}))).
\end{equation}
Both encoders and transformation networks remain frozen at inference.
\section{Experiments}
We present both simulation and real-world experiments. 
Simulation experiments evaluate (1)~whether PFC is an effective representation for tactile information for RL policies and (2)~whether the nearest-neighbor-based rectified flow training effectively bridges the modality gap between PFC and tactile readings, (3) and how \methodabbr~compares to baselines.
Real-world experiments evaluate (1)~how our method compares to physics-based tactile simulation for zero-shot sim-to-real deployment and (2)~whether the approach can be applied across different tactile sensors.
\subsection{Simulation Experiments}
\label{sec:simulation-experiments}
Although our method targets sim-to-real transfer, simulation experiments still provide useful insight.
They aim to transfer a policy from a simulation environment without tactile sensors (the \emph{nominal} environment) to the simulation environment with tactile sensors (the \emph{proxy-real} environment).
The simulation experiments can decouple the modality alignment (PFC $\leftrightarrow$ tactile readings) and dynamics gap of sim-to-real transfer, whereas in the real-world experiments, those two gaps are generally both present to some degree. 
Specifically, we introduce two simulation environments: (1)~a \emph{nominal} simulation environment and (2)~a \emph{proxy-real} environment.
The nominal environment provides only PFC as the tactile representation, with no additional tactile sensor rendering, which is the simulation setting described above. 
The proxy-real environment replaces PFC with physics-based tactile simulation and removes access to PFC. 
It acts as a real-world environment for evaluating transfer, but is still simulated at this simulation experiment stage.
Those two environments share the same dynamics, and are used to evaluate the performance of aligning PFC and tactile readings (modality gap).
Note that physics-based tactile simulation is introduced in simulation experiments NOT because it is a necessary component of our method, but to set up a simulated real-world environment to evaluate our method. We make no claims regarding the accuracy of the physics-based simulation or its suitability for sim-to-real transfer.

\subsubsection{Experiment Setup}
We use Isaac Lab~\cite{mittal2025isaac} to build the simulation environments and focus on three Factory tasks: (1)~Peg Insertion, (2)~Gear Meshing, and (3)~Nut Threading. 
In these Factory~\cite{narang2022factory} tasks, the robot starts with the object already grasped and controls only the arm, not the gripper, to complete the task. 
All three environments randomize the object's initial pose relative to the gripper, making tactile sensing necessary to complete the task. They also randomize the initial robot pose and the pose of the fixed counterpart (e.g., the socket for peg insertion, the gear base for gear meshing, and the bolt for nut threading).

We use TacSL~\cite{akinola2025tacsl} for tactile simulation for setting up the proxy-real environment, assuming a visuotactile sensor and rendering tactile images from interpenetration between the fingertip and the object.

We consider three baselines: (1)~\textbf{Discrete Contact Position (DCP)}. To avoid setting up tactile sensor simulation, Qi et al.~\cite{qi2023general} reduce real tactile readings to information that is also easy to obtain in simulation. Specifically, they use a discretized contact position: in simulation the contact position is directly available; in the real world, for a visuotactile sensor, they track the center of the contact area. Positions are discretized into eight bins in both domains to further mitigate the modality gap. We use this baseline to test whether a richer PFC representation is necessary for sim-to-real transfer. 
(2)~\textbf{Optimal Transport (OT)}. Rather than nearest-neighbor search and rectified flow, Cheng et al.~\cite{cheng2026generalizable} use optimal transport for modality alignment. Although their method was designed to align simulated and real camera images, we slightly modify their method to adapt it to aligning PFC and tactile readings. 
The alignment encoder $f_{\phi}$, originally used to align raw camera images, takes $\mathbf{z}_{p}$ and $\mathbf{z}_{c}$ as input. To avoid producing a degenerate $f_{\phi}$, we apply it only to map $\mathbf{z}_{c}$ to $z\mathbf{z}_{p}$. At inference, actions are sampled as $\mathbf{a} = \pi_{\theta}(\mathbf{o}, f_{\phi}(f_c(\mathbf{c})))$. This baseline tests whether nearest-neighbor search and rectified flow are necessary for modality alignment. (3)~\textbf{No Tactile}. The policy receives only proprioception $\mathbf{o}$, isolating the contribution of tactile sensing.

In addition, we include an ablation, \textbf{PFC + RF}, which removes the second RL stage of Sec.~\ref{sec:rl-training-with-double-transformation}, to show how the information gap between PFC and tactile readings affects policy performance. We also report two oracles: \textbf{Image Oracle} trains directly on tactile images in the proxy-real environment, using the same frozen tactile MAE encoder $f_c$ as our method; it represents the best performance achievable by RL with direct access to the proxy-real tactile readings. \textbf{PFC Oracle} is the policy trained in Sec.~\ref{sec:rl-training-with-pfc} with PFC input and evaluated with PFC input. This is introduced to show performance drop for adapting to tactile sensor inputs. 

All policies are trained with PPO~\cite{schulman2017proximal}. For SBLR, Stage~1 uses 150 PPO iterations (2.46M environment interactions), while Stage~2 uses 20 (0.33M; 13.3\% of Stage~1).

To emphasize the role of tactile information, the observation $\mathbf{o}$ in simulation experiments includes only the robot end-effector pose. Additional observations could improve performance, but a minimal proprioceptive input better highlights the effect of tactile sensing and alignment quality.

For alignment data for each task, we collect 512 episodes of simulation data with PFC and 4096 episodes of proxy-real data with tactile readings.

Our PFC encoder $f_{pfc}$ and tactile encoder $f_{c}$ are trained with data across all three tasks combined together, while RF transformation networks are trained per task. This is to better match the training and inference distribution.
\subsection{Experiment Results}
Table~\ref{tab:sim_results} reports zero-shot success rates in the proxy-real environment across the three Factory tasks. For each method and task, we report the average and maximum success rate over three seeds. Each experiment is evaluated for 256 episodes.

\begin{table}[t]
\setlength{\tabcolsep}{2pt} 
\centering
\vspace{0.3cm}
\caption{Zero-shot success rate (\%) in the proxy-real environment (256 episodes).}
\vspace{-0.3cm}
\label{tab:sim_results}
\begin{tabular}{lcccccc}
\toprule
\multirow{2}{*}{Method} &
\multicolumn{2}{c}{Peg Insertion} &
\multicolumn{2}{c}{Gear Meshing} &
\multicolumn{2}{c}{Nut Threading} \\
\cmidrule(lr){2-3} \cmidrule(lr){4-5} \cmidrule(lr){6-7}
 & Avg & Max & Avg & Max & Avg & Max \\
\midrule
\methodabbr~(Ours) $\uparrow$& \textbf{87.4\%} & \textbf{93.0\%} & \textbf{86.5\%} & \textbf{91.0\%} & \textbf{75.9\%} & \textbf{80.9\%} \\
DCP~\cite{qi2023general} $\uparrow$& 22.5\% & 28.1\% & 25.1\% & 56.3\% & 35.7\% & 43.0\% \\
OT~\cite{cheng2026generalizable} $\uparrow$& 72.9\% & 79.6\% & 75.4\% & 89.1\% & 52.3\% & 55.1\% \\
No Tactile $\uparrow$& 49.2\% & 57.4\% & 32.0\% & 44.5\% & 37.5\% & 49.2\% \\

Ablation~(PFC + RF) $\uparrow$& 80.3\% & 88.7\% & 83.3\% & \textbf{91.0\%} & 48.2\% & 56.6\% \\
\midrule
Image Oracle $\uparrow$& 86.6\% & 88.3\% & 88.2\% & 90.6\% & 87.6\% & 93.8\% \\
PFC Oracle $\uparrow$& 88.5\% & 90.6\% & 88.2\% & 94.1\% & 93.4\% & 94.1\% \\
Image Oracle - \methodabbr $\downarrow$& -0.8\% & -4.7\% & 1.7\% & -0.4\% & 11.7\% & 12.9\% \\
\bottomrule
\end{tabular}
\vspace{-0.5cm}
\end{table}
\methodabbr~consistently improves over its ablation.
The gap relative to the ablation is modest on Peg Insertion and Gear Meshing, likely because the visuotactile sensor in the proxy-real environment already provides much of the high-precision information available in PFC. 
The gap is larger on Nut Threading, where precise alignment between PFC and tactile readings is more critical. The second stage of RL training with bottlenecked latent reconstruction not only adapts to the information gap, but can be potentially used to adapt, to some extent, to alignment error in $f_{c \rightarrow p}(\mathbf{z}_{c})$. Such adaptation is also valuable in real-world experiments, where transformation networks are trained on noisier data.
DCP does not provide sufficiently-precise tactile information for the Factory tasks, and the discrete contact signals create an additional challenge for RL training.
Despite differences in formulation, OT-based pairing between PFC and tactile readings has very similar results to our nearest-neighbor search, and its performance is accordingly close to that of the ablation.
\looseness=-1
\subsection{Real-world Experiments}
\begin{figure}[t]
    \centering
    \includegraphics[width=\columnwidth]{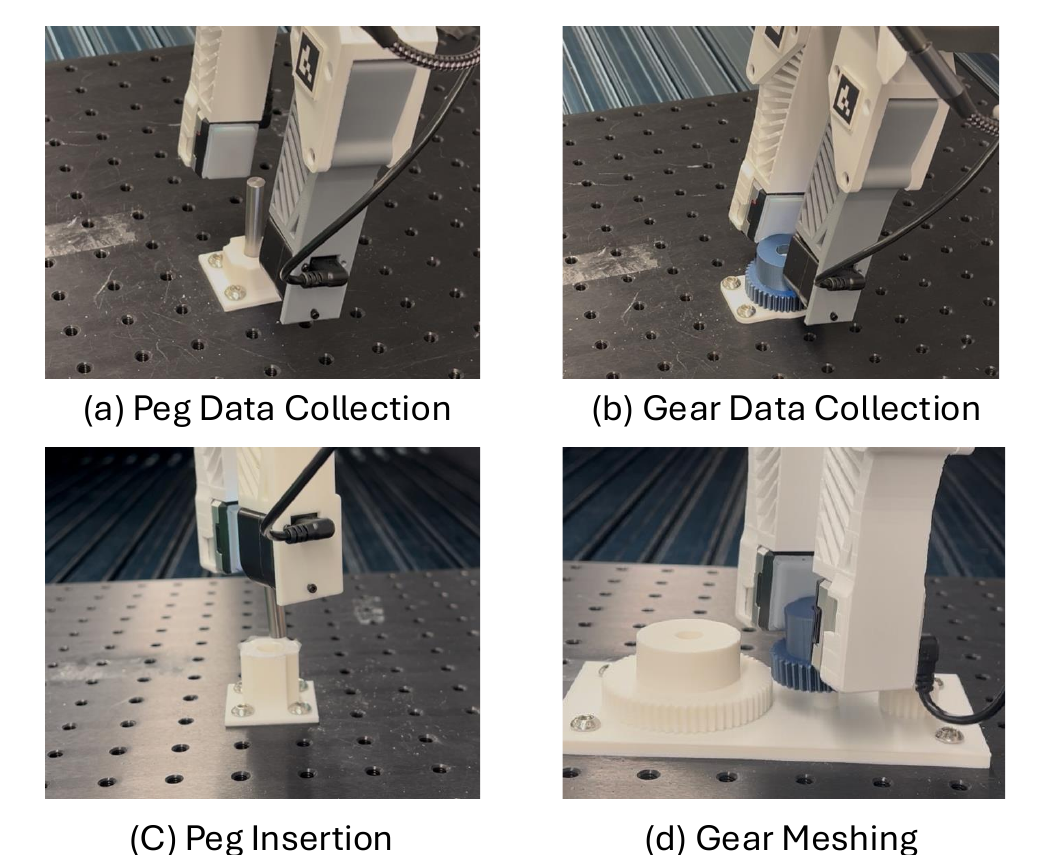}
    \vspace{-0.7cm}
    \caption{Real-world hardware setup. (a) and (b) shows the setup for data collection. (c) and (d) shows the setup for the policy deployment.}
    \label{fig:hardware_setup}
    \vspace{-0.5cm}
\end{figure}
\subsubsection{Experiment Setup}

The hardware experiments deploy the policies trained in the simulation experiments in Sec.~\ref{sec:simulation-experiments} to the real world.
We 3D-print the simulation assets for Peg Insertion and Gear Meshing from the public IsaacLab repository and set up corresponding hardware experiments. 
Relative to the simulation experiments, the RL observation includes an additional previous action, and we randomize robot joint friction, object friction, joint control stiffness, and sensor elastomer stiffness to adapt the policy to the dynamics gap between simulation and the real world.

We compare against TacSL~\cite{akinola2025tacsl} as a physics-based tactile simulation baseline. Specifically, we 3D-print a poking sphere and calibrate the tactile sensor with Taxim~\cite{si2022taxim}. Even after calibration, rendered tactile images in simulation still differ slightly from real readings, so we followed TacSL to apply image augmentation to tactile images during RL training, which includes jittering the brightness, contrast, hue, and slightly cropping the image. TacSL tests whether our method can match physics-based approaches that require careful calibration and simulation setup.
For the TacSL policy, we do not reuse the frozen MAE encoder $f_c$ from our method. Instead, we train a convolutional neural network (CNN) encoder end-to-end with RL, which gives the policy more freedom to extract whatever information from the raw tactile images is useful for the task. Because end-to-end RL from high-dimensional images is difficult to optimize, we split this training into two stages that act as a curriculum: (1)~we first train the policy with tactile image augmentation only, and (2)~we then fine-tune it with dynamics randomization, such as joint friction and control stiffness. Both stages are trained to convergence. 

To show that our method applies across different sensors, we also evaluate on GelSight Mini and DIGIT~\cite{lambeta2020digit}. Both sensors are visuotactile sensors. Compared with GelSight Mini, DIGIT has lower resolution and a stiffer elastomer, yielding tactile images with less detail.

For alignment data for each task and each sensor, we collect 256 real-world episodes with tactile readings and 4096 simulation episodes with PFC. Real-world collection is fully automatic, requiring no human intervention, and takes about 40 minutes. Note that, as in Sec.~\ref{sec:alignment-data-construction}, this is random-play data with a fixed object, not on-task data such as teleoperated expert demonstrations.

We only use the tactile sensor on the left fingertip rather than both sensors, which is sufficient for our tasks. 
% This is because the tactile elastomer on the sensor often tears after extensive use. Using one sensor reduces the number of sensor replacements needed.

\subsubsection{Real-World Experiment Results}
Each experiment is evaluated over 40 episodes. Results are shown in Table~\ref{tab:real_results}.

\begin{table}[t]
\centering
\vspace{0.3cm}
\caption{Zero-shot success rate (\%) on hardware (40 episodes).}
\vspace{-0.3cm}
\label{tab:real_results}
\footnotesize
\begin{tabular}{lcccc}
\toprule
\multirow{2}{*}{Method} &
\multicolumn{2}{c}{GelSight Mini} &
\multicolumn{2}{c}{DIGIT} \\
\cmidrule(lr){2-3} \cmidrule(lr){4-5}
 & Peg & Gear & Peg & Gear \\
\midrule
\methodabbr~(Ours) $\uparrow$ & \textbf{92.5\%} & \textbf{97.5\%} & \textbf{92.5\%} & \textbf{85.0\%} \\
Ablation~(PFC + RF) $\uparrow$ & 85.0\% & 90.0\% & 75.0\% & 75.0\% \\
TacSL~\cite{akinola2025tacsl} $\uparrow$ & 77.5\%  & 82.5\% & 77.5\% & 77.5\% \\
\bottomrule
\end{tabular}
\vspace{-0.5cm}
\end{table}
\methodabbr~consistently improves over its ablation. The gap between our method and the ablation is slightly larger on DIGIT than on GelSight Mini, likely because DIGIT provides less precise tactile information, making the second RL stage of adapting to information gap more important.

TacSL achieves reasonable performance but falls short of our method. Even with physics-based modeling, the simulated tactile image differs from real readings because the model relies on assumptions that do not always hold in practice, for example, that the sensor surface is flat and that tactile images depend only on fingertip--object interpenetration. In addition, we calibrate with a 3D-printed poking sphere.  Due to the 3D printing error, the sphere is not perfectly smooth, and the size is not exactly the same as the mesh, which introduces further calibration error. Calibration could be improved with a higher-quality object, but our setup reflects what most lab-scale hardware can readily achieve. 
Overall, our method outperforms TacSL in our experiments because we train directly on real tactile data from contacts representative of inference-time interactions. This avoids physics-based modeling assumptions and reduces the distribution shift between alignment contact data and deployment contact.
\looseness=-1

% However, we do not claim that our approach is fundamentally superior to physics-based tactile simulation. Accurately simulating tactile sensors remains an active research area, and with sufficiently faithful models, such methods can potentially achieve strong performance.
\section{Discussion and Limitations}
Although our method shows promising results, several limitations remain. (1)~We address the modality gap between simulated PFC and real tactile readings, but not the dynamics gap of sim-to-real transfer. (2)~For high-precision tasks such as nut threading, random-interaction alignment data may not efficiently cover contacts that matter at inference; human teleoperation could potentially improve alignment data quality and coverage.

Our framework may also have broader applications beyond aligning an oracle sensor in simulation and real sensors for sim-to-real deployment: it offers a unified way to adapt a pretrained policy to different sensors at deployment. 
For example, a large pretrained model (e.g., a vision-language-action model) with a certain sensor can be quickly adapted to new sensor setups via Stage~2 RL, avoiding expensive full retraining for each sensor.
\section{Conclusion}
We presented \methodabbr, a framework for zero-shot sim-to-real transfer that trains policies on a simulator-native oracle sensor and aligns real sensor latents to the oracle latents at inference via unpaired random-play data, avoiding sensor-specific simulation. A lightweight second-stage RL procedure with bottlenecked latent reconstruction further adapts the policy to the information gap between the oracle sensor latents and real sensor latents. We instantiate the framework on tactile sensing with Point Force Composition (PFC) as the oracle sensor reading. Simulation and hardware results on Factory tasks with GelSight Mini and DIGIT show that this approach is an effective alternative when sensor-specific simulation is unavailable or costly to set up.
\appendices
\section{Alignment Data Collection and Pseudo-Pairing}
\label{app:alignment-data}
\subsection{Data Collection}
\label{sec:alignment-data-collection}
We fix the object to be manipulated in the task (e.g., the peg in peg insertion, or the gear in gear meshing) and collect data by commanding the robot to make random probing contacts with the fixed object.
To do this we randomly sample an end-effector pose centered on the fixed object, move to this pose, and close the gripper to establish contact. We then apply a small random probing action that slightly translates and rotates the end effector, approximating the sliding and rotational contacts that arise during policy inference.

We use a fixed object for two reasons: (1)~a fixed object provides a known object pose. Together with robot proprioception, it defines the robot-object contact and thus supports pseudo pairing for contact; and (2)~random interaction with a fixed object requires no additional human effort such as manual object resetting or teleoperation, and can be fully automated on hardware.
In simulation, full point clouds in PFC are obtained by reading object and robot meshes and rendering points from the current configurations.
\subsection{Pseudo-Pairing}
\label{sec:pseudo-pairing}
We construct pseudo-pairs with nearest-neighbor search over proprioceptive transitions:
\begin{equation}
S_t^s := (\mathbf{q}_{t-1}^s, \mathbf{q}_{t}^s), \quad S_t^r := (\mathbf{q}_{t-1}^r, \mathbf{q}_{t}^r),
\end{equation}
where superscripts $s$ and $r$ denote simulation and the real world respectively. Similarity between an oracle reading $\mathbf{p}_t$ and tactile reading $\mathbf{c}_t$ is defined based on the difference of its corresponding proprioceptions:
\begin{equation}
        D(S_t^s, S_t^r) = ||\mathbf{q}_t^s - \mathbf{q}_t^r||_2 + \lambda||\mathbf{q}_{t-1}^s - \mathbf{q}_{t-1}^r||_2,
\end{equation}
where $\lambda < 1$ is a weighting parameter. Only robot proprioception is used, with no object information, because the object is fixed at the same pose in both domains during data collection.
We exploit the slight deformation of the tactile elastomer under contact so that normal force is reflected in proprioception: a larger normal force brings the fingertip closer to the object. Shear force, however, is not directly reflected in a single proprioceptive state; including the previous time step helps capture contact dynamics through the proprioceptive transition.
Robot actions are not used for the similarity computation, since simulation and the real system typically have different dynamics (e.g., joint friction), so the same action command may not produce the same joint actuation.
\looseness=-1

For each real-world sample, since some samples might not have their pairs close enough from the simulation dataset, we first retain simulation samples that satisfy $D(S_t^s, S_t^r) < \delta$ for a predefined threshold $\delta$.
Among the retained candidates, we then select the top-$k$ nearest simulation samples as pseudo-pairs.
\bibliographystyle{IEEEtran}
\bibliography{reference}

%                                   % the textheight of the last page by a suitable amount.
%                                   % This command does not take effect until the next page
%                                   % so it should come on the page before the last. Make
%                                   % sure that you do not shorten the textheight too much.

\end{document}